\documentclass[runningheads]{llncs}
\usepackage[T1]{fontenc}
\usepackage{graphicx}
\usepackage{multirow}
\usepackage{outlines}
\begin{document}
\title{GoNoGo: An Efficient LLM-based Multi-Agent System for Streamlining Automotive Software Release Decision-Making}

\titlerunning{GoNoGo: Software Release Multi-agent System}
%
\author{Arsham Gholamzadeh Khoee\inst{1,2}\orcidID{0000-0002-5130-5520} \and
Yinan Yu\inst{1}\orcidID{0000-0002-3221-7517} \and
Robert Feldt\inst{1}\orcidID{0000-0002-5179-4205} \and
Andris Freimanis\inst{2} \and
Patrick Andersson Rhodin\inst{2} \and
Dhasarathy Parthasarathy\inst{2}\orcidID{0000-0002-3620-8589}
}
\authorrunning{A. Khoee et al.}
%

\institute{Department of Computer Science and Engineering, Chalmers University of Technology, Gothenburg, Sweden 
\email{\{khoee, yinan, robert.feldt\}@chalmers.se}\\ \and
Volvo Group, Gothenburg, Sweden
\email{\{andris.freimanis, patrick.andersson, dhasarathy.parthasarathy\}@volvo.com}}
\maketitle              
\begin{abstract}
Traditional methods for making software deployment decisions in the automotive industry typically rely on manual analysis of tabular software test data. 
These methods often lead to higher costs and delays in the software release cycle due to their labor-intensive nature. Large Language Models (LLMs) present a promising solution to these challenges. However, their application generally demands multiple rounds of human-driven prompt engineering, which limits their practical deployment, particularly for industrial end-users who need reliable and efficient results. 
In this paper, we propose GoNoGo, an LLM agent system designed to streamline automotive software deployment while meeting both functional requirements and practical industrial constraints. Unlike previous systems, GoNoGo is specifically tailored to address domain-specific and risk-sensitive systems. We evaluate GoNoGo's performance across different task difficulties using zero-shot and few-shot examples taken from industrial practice. Our results show that GoNoGo achieves a 100\% success rate for tasks up to Level 2 difficulty with 3-shot examples, and maintains high performance even for more complex tasks. We find that GoNoGo effectively automates decision-making for simpler tasks, significantly reducing the need for manual intervention. 
In summary, GoNoGo represents an efficient and user-friendly LLM-based solution currently employed in our industrial partner's company to assist with software release decision-making, supporting more informed and timely decisions in the release process for risk-sensitive vehicle systems.

\end{abstract}
\section{Introduction}
\label{sec:intro}

In the automotive industry, decisions about when to release software, particularly embedded software in risk-sensitive systems, carry immense weight. The complexity of modern vehicles, with their multiple levels of integration, further complicates this process. Each integration level involves one or more gating steps, with tests conducted to verify whether gate criteria are fulfilled. Gate failures can delay the integration of all dependent subsystems, regardless of their individual quality. In this intricate process, release managers, bearing the responsibility of gatekeeping could greatly benefit from assistance to make faster and better decisions.

\begin{figure}[ht]
\centering
\includegraphics[width=0.7\linewidth]{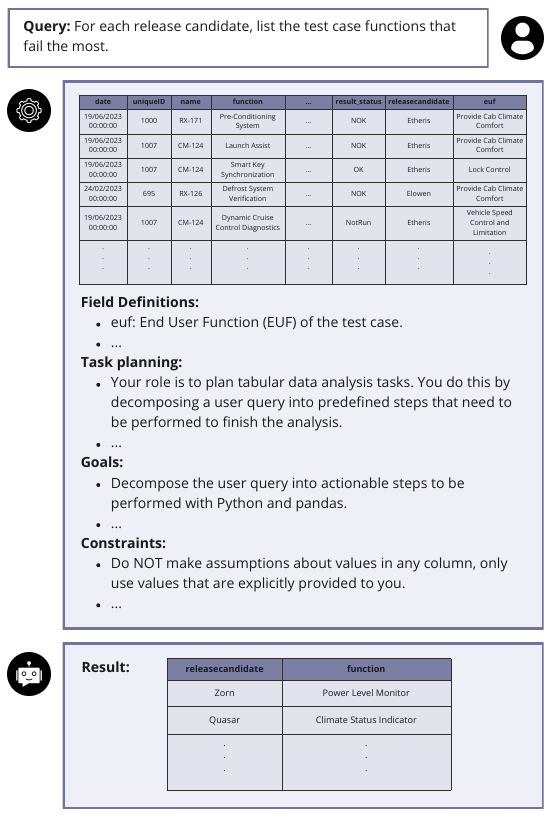}
\caption{An actual example demonstrating the use of the LLM-based multi-agent system for automating ad-hoc tabular data analysis.}
\label{fig:chat}
\end{figure}

Large language models (LLMs) present an interesting avenue for providing such assistance. In particular, LLMs have demonstrated strong capabilities in zero- and few-shot settings with in-context learning~\cite{brown2020language}. Recent advancements have improved reasoning~\cite{zhang2023igniting}, exemplar selection, and prompt design~\cite{chang2024survey}. Companies now use LLMs for software engineering tasks like API testing, code generation, and documentation and research studies have already shown test automation improvements over the state-of-the-art~\cite{yoon2024llmagenttesting}. 

However, when applying LLMs to \emph{risk-sensitive} \emph{domain-specific} tasks, several unique challenges must be addressed. In our research with industrial partners, the most prominent challenges include 1) Incorporating specific logic and terminology relevant to the domain; 2) Understanding and parsing high-level queries or vague language used by non-expert stakeholders and translating them into actionable plans, 3) Enabling interpretability so that domain experts can explain system functionality to stakeholders without excessive complexity, 4) Operating efficiently to meet the time-critical demands of organizational applications, mitigating potential bottlenecks related to limited LLM licensing or infrastructure, and 5) Designing the system to enable ease of troubleshooting and maintenance, ensuring that any issues can be quickly identified and resolved. 


To address these challenges, we propose a multi-agent system that encodes domain-specific requirements using in-context learning. This system comprises two primary LLM agents: a Planner and an Actor (refer to Figure~\ref{fig:system}). The Planner, which forms the core of the system, comprehends and decomposes user queries into step-by-step instructions for data analysis~\cite{khot2022decomposed}. The Actor then synthesizes and generates executable scripts from these higher-level instructions. Within the Actor, a coder LLM utilizes the self-reflection mechanism besides a memory to produce the most effective Python script optimized for querying the given data for each instruction generated by the Planner~\cite{austin2021program,yoon2024llmagenttesting}. 

This system provides an interface for end-users at our industrial partner, illustrated by Figure~\ref{fig:chat}, which shows a real-world example of its use. It allows end-users, like release managers, to interpret results from a business and safety perspective without needing detailed technical knowledge. For example, they can simply receive a short table that reports the test case functions that have failed the most for each release candidate as shown in Figure~\ref{fig:chat}. By reviewing this information, release managers can make well-informed decisions on whether to release the software or not, ensuring that it meets both business objectives and safety standards in the automotive industry. This approach can significantly reduce time and resources by eliminating the need for various database and programming experts to achieve the desired results for end-users. Our agent automates test data analysis across multiple vehicle development integration levels, providing detailed reports on component functionality and system interactions. This assists release managers in making informed decisions about software readiness for release, accelerating development while enhancing gatekeeping reliability. Our contributions can be summarized as follows:

\begin{outline}
 \1 We highlight the practicality of the proposed LLM-based intelligent assistant in making software release decisions within the automotive industry. This is achieved by enhancing two key capabilities: 
   \2 \textbf{Domain-specificity:} We design a framework to handle unstructured queries from non-expert stakeholders in the automotive industry by mapping generic language to domain-specific logic using in-context learning.
   \2 \textbf{Risk-sensitivity:} We incorporate two predefined atomic operations to restrict the action space and improve the risk-sensitive aspect of the planner. 
 \1 Experiments on a total of 50 crafted test queries show that our proposed system is effective at analyzing data and deriving the required insights for software release decision-making.
 \1 Our system, now deployed and actively used within our industrial partner's company, has demonstrated significant improvements in the software release decision-making process besides saving time, improving accessibility, reducing reliance on specialized analysts, and accelerating overall workflow.
\end{outline}

The remainder of this paper is structured as follows: Section \ref{sec:release} provides an overview of the manual process behind automotive software release decisions and the need for streamlining operations. Section \ref{sec:method} details our approach, including a description of the architecture of our LLM-based multi-agent system and an explanation of the Planner and Actor agents. Furthermore, in Section \ref{sec:experiments}, we present our experimental setup and results. Section \ref{sec:related_work} provides an overview of similar research in LLMs for data analysis. Finally, Section \ref{sec:conclusion} concludes the paper by summarizing key findings and discussing the broader implications of our work in the context of industrial software release management and risk-sensitive systems.

\section{Manual Process of Release Decisions: Insights From the Industry}
\label{sec:release}
Deciding to go ahead, or not, with a software release in the automotive industry is a complex task involving multiple stakeholders and extensive data analysis. This section reviews the current, and typical of the industry at large, manual workflow and the need for streamlining.

Vehicle development progresses through multiple phases, each becoming more complex as more components are integrated. Numerous tests are conducted at each stage to ensure functionality and identify revisions, generating vast amounts of data. Software components require repeated testing and validation, adding to this data.

Project managers, verification engineers, and quality engineers need clear analytics and insights from these tests in order to make software release decisions. Extracting essential information is time-consuming. Quality engineers analyze data for continuous improvement, while release engineers need specific information to make informed release decisions.

Within this process, statisticians provide an overall view of the data to project managers and quality engineers for future business decisions. Manual data processing is necessary due to the critical nature of these decisions and their impact on consumer safety. However, this approach is time-consuming and prone to errors, partly due to the differing perspectives of technical data analyzers and statisticians, who may not fully understand the project managers' goals.

A critical and typical stage in this process is ``Testing on Closed Track,'' where vehicles equipped with the necessary software release undergo systematic and rigorous testing of their systems in a controlled environment. After these tests, release managers analyze large amounts of data to decide whether to move to the next test stage. This involves manually querying data to generate reports that support informed decisions. Errors or delays in this analysis can hinder timely software release, affect business goals, and delay subsystem integration.

The deployment of an intelligent assistant has the potential to facilitate software release decisions in the automotive industry~\cite{wang2024surveyl}, particularly during the critical testing on a closed track phase. In this work, we have focused on designing such an LLM-based multi-agent system to address the challenges of this specific stage. By rapidly processing test data from closed track testing, the system can generate comprehensive reports tailored to different stakeholders' needs. For example, it can quickly compile summaries of failed tests, highlight software performance trends across vehicle models, or analyze a specific component's behavior under various conditions. Consequently, this reduces the time spent on initial analysis, allowing release managers to focus on interpreting results and making informed decisions. This not only accelerates the development process but also enhances the accuracy and reliability of the information used in release decisions, ultimately contributing to maintaining high safety and quality standards in automotive software development.

\section{GoNoGo: Intelligent Software Release Assistant}
\label{sec:method}

\subsection{System Requirements}
After discussing current needs and opinions about the software release analysis and decision-making processes with our industrial partner, we identified the following main challenges in automating data analysis:
\begin{description}
   \item[Understanding User Queries] The system must interpret queries, typically presented in natural language~\cite{rahimi2024integrating}, within the specific domain context, using any provided domain-specific knowledge.
   \item[Translating User Queries to Actionable Steps] The system needs to convert the user's query into concrete steps, breaking down complex queries into simpler tasks, determining the order of operations, and selecting appropriate data manipulation or analysis techniques. Additionally, the action space must be carefully managed to adhere to risk-sensitive requirements.
   \item[Execution and Result Preparation] The system must execute the planned actions, interact with data using scripts (e.g., querying databases, performing calculations, applying filters), and compile the results into the desired format for the user.
\end{description}
These steps rely heavily on the LLM's domain-specific knowledge and reasoning ability, crucial for effective query instruction planning~\cite{valmeekam2023planning}. Consequently, this work explores techniques for enhancing the reasoning capabilities of LLM agent systems, particularly for the analysis of tabular data in industrial contexts.

\subsection{System Architecture}
\begin{figure*}[ht]
\centering
\includegraphics[width=1\linewidth]{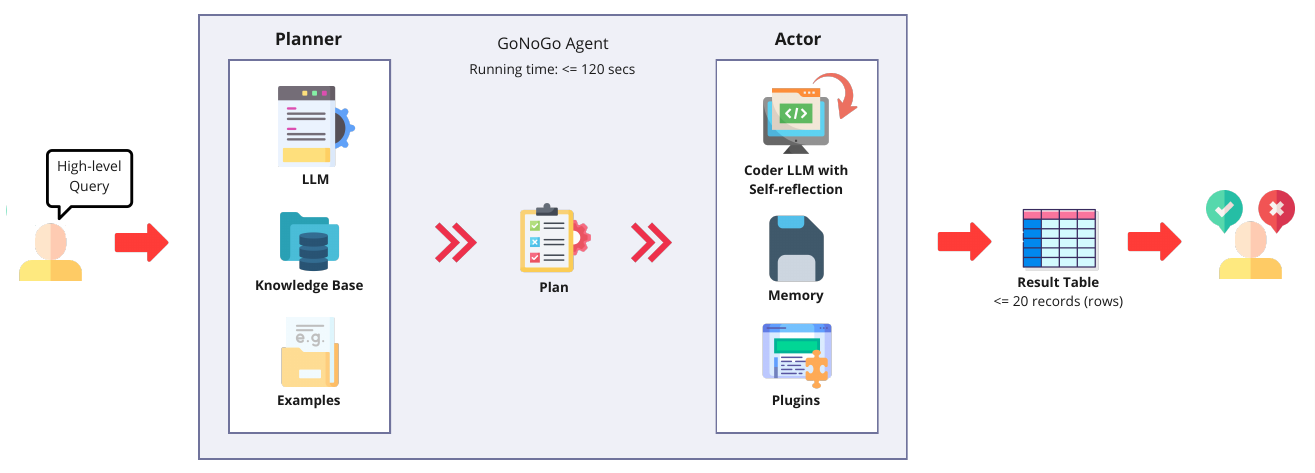} 
\caption{Architecture of the LLM-based multi-agent system GoNoGo along with the illustration of the interaction procedure of the system. GoNoGo receives high-level queries from the end user, performs the required data manipulations, and outputs the result table as a decision support resource. GoNoGo comprises a Planner agent, which interprets queries and devises analysis strategies using Chain-of-Thought prompting and self-consistency, supported by a \emph{Knowledge Base} and \emph{Examples} for few-shot learning. The Actor includes a Coder LLM with a \emph{Self-reflection} mechanism, utilizing \emph{Memory} and \emph{Plugins} for code generation and error resolution. The total running time of GoNoGo for one user query is approximately 120 seconds, which satisfies typical user requirements.}
\label{fig:system}
\end{figure*}

Our approach to automating tabular data analysis leverages LLMs to create an intelligent system capable of interpreting natural language queries, executing complex analyses, and delivering desired results. The system architecture consists of two main components: the Planner supported by a Knowledge Base and Examples for few-shot learning and the Actor including coder LLM, memory module, and some Plugin components. Figure~\ref{fig:system} illustrates the overall architecture of the developed system.

\subsubsection{Planner}

The Planner is the core of our system, responsible for interpreting user queries and devising appropriate analysis strategies. One of the core challenges of designing an LLM-based multi-agent system is the inherent inaccuracy of prompting. As decision-making becomes more distributed over multiple LLM agents, the uncertainty within the multi-agent system increases. To mitigate this, we centralize the complexity within the Planner, which is responsible for the majority of design choices. By focusing on the Planner as the main agent for refinement, we aim to create a system that is both interpretable and easily maintainable. 

\begin{figure}[ht]
\centering
\includegraphics[width=0.8\linewidth]{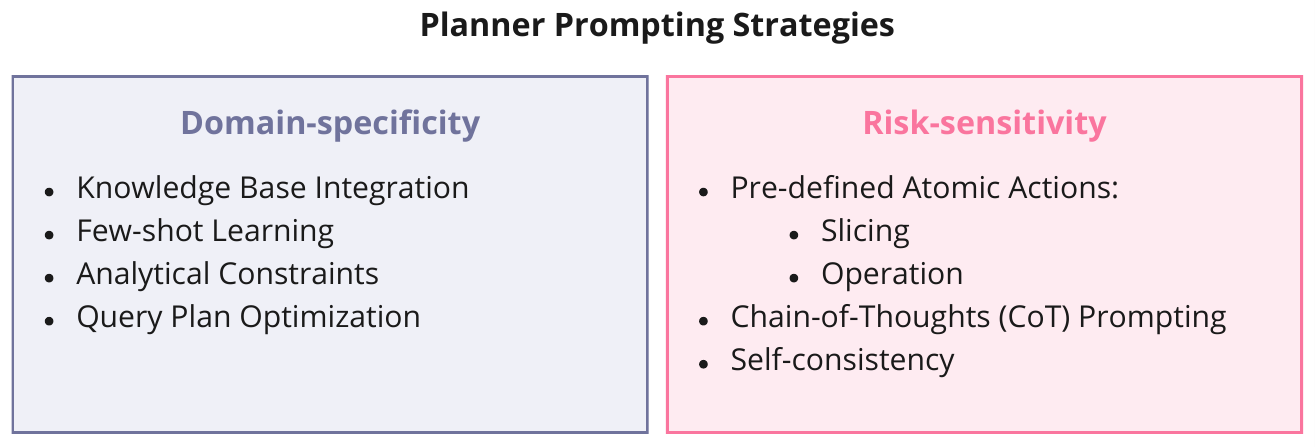}
\caption{Planner prompting strategies addressing domain-specificity and risk-sensitivity in the LLM-based agent system for tabular data analysis.}
\label{fig:planner}
\end{figure}

Our problem consists of two main aspects: domain-specificity and risk-sensitivity. These two characteristics frequently manifest together in real-world applications, particularly in fields such as healthcare and automotive, where unreliability and inaccuracies can have significant consequences. However, there is a noticeable gap in addressing both aspects simultaneously, let alone demonstrating such systems in practice. As part of our system, we want to explicitly address both of these aspects. As the Planner is the component with the most decision-making responsibility, these two requirements are encoded into the Planner prompts as depicted in Figure~\ref{fig:planner}.

\subsubsection{Domain-specificity} 
The Planner utilizes a Knowledge Base containing a structured description of the data and its attributes to provide the necessary context and domain-specific information in the prompts given to the LLM, enhancing the system's performance and applicability \cite{liu2023pre}. The Knowledge Base serves as a comprehensive repository of metadata, including detailed descriptions of data tables, possible states and values for essential fields, and domain-specific terminologies, as well as the semantic meanings of various data elements. This enables the system to understand and interpret high-level queries and devise appropriate analysis plans by using this information as input prompts for the Planner, taking advantage of in-context learning. This integration ensures that the entire pipeline, from query interpretation to result generation, is informed by relevant domain knowledge, enabling the LLM agent to provide more accurate, relevant, and specialized responses to user queries.

In our system architecture, we also feed some input-output pairs as examples into the Planner, allowing few-shot learning alongside the Knowledge Base. This combination enables the Planner to interpret user queries more effectively, drawing on both general knowledge and specific task examples to formulate appropriate analysis plans. This approach makes the system a powerful tool for automated tabular data analysis across various industries and use cases.

Helpful prompts serve as constraints, enhancing the LLM's reasoning capabilities~\cite{huang2022towards}. For example, constraints help the model understand that queries should account for more than just binary states for some fields. Retrieving records with `A' and its opposite doesn't always mean retrieving all records, as other non-binary states might exist. For instance, `successful' and `failed' tests don't encompass all possible test statuses; there may be additional statuses to consider, such as `N/A', that the model should take into account.

The focus is on pushing the model to generate an optimized query plan. This involves narrowing down data through filtering and selection before performing sorting and other operations on the reduced dataset to minimize processing. Accordingly, designed constraints help the agent explore the characteristics of each field and the data, providing more accurate planning. 

\subsubsection{Risk-sensitivity}
We guide the Planner with two pre-defined atomic actions to limit the action space of the planner: slicing and operation. Slicing involves specifying the columns to select and the conditions for filtering rows from the data to be analyzed. Operation involves describing the operations (such as max, mean, count, etc.) to be performed on the values of one or more columns of the data obtained from the slicing step. The steps should be returned as a Python list, with each step described in natural language, including all relevant values and column names.

Also, we leverage Chain-of-Thought (CoT) prompting to further enhance the reasoning capabilities of our LLM-based agent~\cite{wei2022chain}. This technique incorporates intermediate reasoning steps into the prompt, guiding the model to break down complex problems into smaller, more manageable steps~\cite{zhou2022least}. This approach mimics human-like reasoning and problem-solving processes. Additionally, CoT prompting makes the agent's decision-making process more transparent by explicitly showing the reasoning steps, allowing users to understand how the agent arrived at a particular conclusion or analysis result.

We combine CoT prompting with few-shot learning by providing examples that not only show input-output pairs but also include the intermediate reasoning steps. This synergy further enhances the agent's ability to handle diverse and complex data analysis tasks~\cite{dagdelen2024structured}.

To further improve reasoning, we employ self-consistency in conjunction with CoT prompting. This involves generating multiple independent reasoning paths for the same query, comparing them for consistency, and using majority voting to determine the most reliable outcome~\cite{wang2022self}. By considering multiple reasoning paths, the system becomes less likely to be misled by a single flawed chain of thought. As a result, for queries with potential ambiguity, self-consistency can help identify different valid interpretations and provide a more comprehensive answer.

\subsubsection{Actor}

The Actor is responsible for carrying out the analysis plans devised by the Planner. It consists of several interacting components: Coder LLM with Self-reflection, Memory, and Plugins, as depicted in Figure~\ref{fig:system}.

The Coder LLM is responsible for generating executable scripts based on the Planner's instructions. This component is crucial as it translates abstract plans into concrete, executable code that can interact with the data using the required Plugins and perform the necessary analysis. It includes a Self-reflection mechanism, which works in tandem with a Memory module. This Memory stores generated code, error messages, execution results, and contextual information about the current task~\cite{chen2023teaching}.

The Self-reflection mechanism is a sophisticated process that allows the Coder LLM to critically analyze its own output and decision-making process. When an error occurs during script execution, the Self-reflection mechanism activates, providing feedback to the Coder LLM. This feedback loop enables the LLM to analyze error messages within the task context~\cite{dyachenko2018approaches,yoon2024llmagenttesting}, facilitating iterative improvement of the generated code.

The Self-reflection mechanism offers several advantages: It enables the Coder LLM to autonomously identify and correct errors by continuously analyzing and reflecting on its own output, thereby reducing the need for external debugging and intervention. This mechanism promotes a cycle of continuous improvement, allowing each iteration to refine the scripts for progressively better performance and reliability~\cite{yao2022react}. By utilizing the Memory module, the Coder LLM can make context-aware adjustments, considering previous errors, execution results, and specific task requirements, which leads to more precise and contextually appropriate code generation. Automated error correction and iterative refinement result in a more efficient coding process, speeding up the development cycle and enhancing the robustness and reliability of the final scripts. Additionally, the self-reflective capabilities minimize the need for human intervention in the debugging process, enabling engineers to focus on more complex and high-level tasks.

This architecture enables the Actor to not only generate code for data analysis tasks but also to troubleshoot and improve its own output, resulting in a more robust and reliable automated data analysis system.

\subsection{System Implementation}
The system uses Azure OpenAI's GPT-3.5 Turbo for both the Planner and Actor agents. The Planner utilizes specially designed prompts for task planning, defining the entire data analysis task by specifying the details of each step in the plan. Moreover, the Actor uses predefined prompts to generate the required Python code for executing each step of the provided plan with the pandas library, performing tasks on the given data.

\section{Experiments}
\label{sec:experiments}
\subsection{Data}
The data used for analysis at our industrial partner is called ``GoNoGo'' data and is updated after testing each function of every software component in each vehicle. This internal company data contains about 40 different fields and is critical for release decisions, as it includes detailed information regarding the performance and functionality of software components. It provides the necessary information for determining whether to advance a vehicle to the next phase of development and allow it to be driven on open roads. Although the data is updated after each test, we used a dataset of 55,000 records to report our experiments.

Stakeholders often ask questions like ``What are the test case functions that fail the most for release candidate X?'' or ``What is the Y-status of X?'' where X is the release candidate's name and Y is a specific functionality. Answering these questions requires domain knowledge and an understanding of the data to extract and communicate the answers accurately. By analyzing this data, release managers can determine if a vehicle meets the necessary criteria to progress to the next development phase or be driven on public roads. This ensures that only vehicles that meet stringent safety and quality standards are advanced, maintaining high standards in automotive software development.

\subsection{Benchmark Overview}
To evaluate the GoNoGo system's performance, we developed a benchmark based on 15 initial analysis tasks. These tasks were defined with the help of release engineers, quality engineers, and verification engineers. We identified the most common high-level queries and criteria frequently used by these end users in their workflows. Our goal was to design tasks that capture the nuances and complexity of the demanding queries necessary for their decision-making processes. These tasks were then translated into explicit table analysis queries that GoNoGo could process, ensuring that the benchmark reflects real-world scenarios and challenges typically encountered by these professionals.
We created definitive ground-truth solutions for these queries using Python, breaking down the solutions into smaller code chunks representing operations such as filtering, grouping, and sorting. For each query, we generated a series of query ablations by incrementally adding code chunks and formulating corresponding queries that these chunks would solve. This method expanded our original 15 queries into 50 query ablations, each with a corresponding ground-truth solution and Python code.

In this way, we established queries with four levels of difficulty:

\begin{description}
   \item[Level 1] These are the simplest queries, typically involving a single operation such as filtering or sorting.
   \item[Level 2] These queries combine two or three basic operations, such as multiple filtering followed by sorting.
   \item[Level 3] These queries involve more than three operations, potentially including grouping and aggregating. 
   \item[Level 4] These are the most complex queries, requiring multiple advanced operations such as grouping and aggregating, for calculating statistics, beyond basic filtering and sorting.
\end{description}

This incremental approach to query complexity allows us to assess GoNoGO's performance at various levels of difficulty. It helps identify at which point, if any, the system's performance begins to degrade, and provides insights into its capabilities in handling increasingly complex table analysis tasks.

This benchmark allows for objective evaluation of the GoNoGo's ability to handle increasingly complex table analysis tasks, ensuring a comprehensive assessment of its performance across a spectrum of difficulty levels.

\subsection{Evaluation}
The evaluation process involves comparing the GoNoGo system's results against manually generated ground-truth results. The comparison is based on a strict matching criterion~\cite{chiang2024chatbot}. For a match to be considered successful, the system's output must contain the same columns as the ground truth. Additionally, each record in the system's output must exactly match a corresponding record in the ground truth, including all values across different fields. The system's output must also contain the same number of records as the ground truth, with no missing or extra entries. This strict matching ensures that the output is not just similar, but identical in structure and content to the expected result.
If the agent's output satisfies all these criteria when compared to the ground truth, the task is marked as successful; otherwise, it is considered a failure. The model's performance is then quantified by calculating the success rate, defined as the ratio of successful tasks to the total number of tasks. 

\subsection{Results}

We present our experiment results on the GoNoGo system in Table \ref{table:res}. We evaluated its performance across different levels of task difficulty using 0-shot, 1-shot, 2-shot, and 3-shot examples. GoNoGo achieved high performance with 3-shot examples.

\begin{table*}

  \centering
   \caption{Performance evaluation of the GoNoGo system with varying numbers of example queries across different levels of task difficulty.} 
   \label{table:res}
   \resizebox{1\textwidth}{!}{
  \begin{tabular}{p{2.5cm}p{2.5cm}p{2.5cm}p{2.5cm}p{2.5cm}p{2.5cm}}
    \hline
    \textbf{\# Examples}  & \textbf{Task Difficulty}  & \textbf{\# Total Tasks}   & \textbf{\# Success}   & \textbf{\# Failed} & \textbf{Performance}  \\
    \hline
    \multirow{4}{*}{0-shot} & 1       & 16   & 3  & 13  & 18.75\%        \\
                            & 1-2     & 32   & 6  & 26  & 18.75\%        \\
                            & 1-3     & 44   & 9  & 35  & 20.45\%        \\
                            & 1-4     & 50   & 11  & 39  & 22\%        \\
    \hline
    \multirow{4}{*}{1-shot} & 1       & 16   & 15  & 1  & 93.75\%        \\
                            & 1-2     & 32   & 27  & 5  & 84.37\%        \\
                            & 1-3     & 44   & 32  & 12  & 72.72\%        \\
                            & 1-4     & 50   & 34  & 16  & 68\%        \\
    \hline
    \multirow{4}{*}{\textbf{2-shot}} & \textbf{1}       & \textbf{16}   & \textbf{16}  & \textbf{0}  & \textbf{100\%}        \\
                            & 1-2     & 32   & 31  & 1  & 96.87\%        \\
                            & 1-3     & 44   & 38  & 6  & 86.36\%        \\
                            & 1-4     & 50   & 41  & 9  & 82\%        \\
    \hline
    \multirow{4}{*}{\textbf{3-shot}} & \textbf{1}       & \textbf{16}   & \textbf{16}  & \textbf{0}  & \textbf{100\%}        \\
                            & \textbf{1-2}     & \textbf{32}   & \textbf{32}  & \textbf{0}  & \textbf{100\%}        \\
                            & \textbf{1-3}     & \textbf{44}   & \textbf{41}  & \textbf{3}  & \textbf{93\%}        \\
                            & \textbf{1-4}     & \textbf{50}   & \textbf{45}  & \textbf{5}  & \textbf{90\%}        \\
    \hline
  \end{tabular}
  }
\end{table*}

Initially, we assessed GoNoGo's ability to handle the simplest queries involving basic operations like filtering or sorting (Level 1). We then incrementally increased the complexity by including queries that combined Level 1 and Level 2 difficulties, followed by those incorporating Level 1 to Level 3 difficulties. Finally, we evaluated GoNoGo's performance on the full spectrum of tasks, including the most complex queries (Level 4), which require multiple operations such as filtering, sorting, grouping, and calculating statistics.

Our observations indicate that GoNoGo with 3-shot examples is particularly effective for solving queries with task difficulty up to Level 2 and can handle these tasks without error. For more complex tasks involving Level 3 or Level 4 difficulties, human intervention is recommended to perform the necessary manipulations and computations, rather than relying solely on the automated system.

\section{Threats to Validity}
\label{sec:limitation}
We identify the following threats to the validity of our study:

\begin{description}
\item[Limitation of the Created Benchmark]
Our study relies on a benchmark specifically created for evaluating the system. While this benchmark is designed to be comprehensive, it may not cover all potential scenarios and edge cases encountered in real-world applications. Efforts have been made to design queries and tasks to be as comprehensive as possible by involving verification engineers to mitigate subjectiveness. However, despite these efforts, the limitation remains that it may not capture every potential scenario and edge case. This limitation could affect the generalizability and robustness of our findings.

\item[Selection of the Foundation Model]
The choice of the foundation model, in this case GPT-3.5 Turbo, which is considered a widely used LLM in recent studies, might influence the results. Different foundation models, such as GPT-4, GPT-4o, Claude 3, or LLaMA 3, may yield better performance levels and interpretations of the same tasks. However, we limited the project to using GPT-3.5 Turbo and focused on improving its reasoning and planning capabilities. Besides, our framework is flexible and can be easily applied to different pre-trained models. The dependency on a single model means that our conclusions may not hold if another model were used.
\end{description}

\section{Related Work}
\label{sec:related_work}
The application of tabular data in machine learning holds significant potential, ranging from few-shot learning for data analysis to end-to-end data pipeline automation.

Integrating LLM with tabular data presents several substantial challenges \cite{van_breugel_why_2024}. Most foundation models are not trained on tabular data, making it difficult for them to process and interpret this type of data effectively. To mitigate this issue, pre-training LLMs using tabular data or fine-tuning on specific tasks are two commonly adopted options.  \cite{patil2024review} described different phases and strategies for LLM training, and \cite{vm2024fine} provided guidelines for enterprises who are interested in fine-tuning LLMs.

In particular, recent literature has seen a growing interest in pre-training and self-supervised learning (SSL) approaches using tabular data. \cite{wang2024survey} emphasizes SSL for non-sequential tabular data (SSL4NS-TD), categorizing methods into predictive, contrastive, and hybrid learning, and discussing application issues such as automatic data engineering and cross-table transferability. In contrast, \cite{zhang2023generative} introduces TapTap, a novel table pre-training method that enhances tabular prediction and generates synthetic tables for various applications. Finally, \cite{ye2023training} introduces Tabular data Pre-Training via Meta-representation (TabPTM), which enables training-free generalization across heterogeneous datasets by standardizing data representations through distance to prototypes. The common theme across these works is the enhancement of tabular data handling through innovative pre-training and SSL techniques, though they differ in their specific methodologies and application focuses, ranging from generating synthetic data to improving model generalization and manipulation capabilities. \cite{zhang2023towards} proposes Tabular Foundation Models (TabFMs), leveraging a pre-trained LLM fine-tuned on diverse tabular datasets to excel in instruction-following tasks and efficient learning with scarce data.

Pre-training aims to enhance LLMs' capability of handling tabular data in general. However, it does not necessarily improve their performance on specific tasks. On the other hand, fine-tuning pre-trained LLMs have demonstrated potential for enhancing tabular data manipulation on specific tasks. \cite{zhang_tablellm_2024} introduced TableLLM, a robust 13-billion-parameter model designed for handling tabular data in real-world office scenarios. 
In particular, TableLLM incorporates reasoning process extensions and cross-way validation strategies, outperforming existing general-purpose and tabular-focused LLMs. 
\cite{hegselmann_tabllm_nodate} explored zero-shot and few-shot tabular data classification by prompting LLMs with serialized data and problem descriptions, achieving superior performance over traditional deep-learning methods and even strong baselines like gradient-boosted trees. \cite{zhu_tat-llm_2024} addressed question answering over hybrid tabular and textual data, fine-tuning LLaMA 2 using a step-wise pipeline, resulting in TAT-LLM, which outperforms both prior fine-tuned models and large-scale LLMs such as GPT-4 on specific benchmarks. \cite{yang_unleashing_2024} focused on applying LLMs to predictive tasks in tabular data, enhancing LLM capabilities through extensive training on annotated tables.

\paragraph{Industrial considerations}
One known issue is that LLMs often memorize tabular data verbatim, leading to overfitting. \cite{bordt2024elephants} highlights that despite their nontrivial generalization capability, LLMs perform better on datasets they were exposed to during training compared to new, unseen datasets. This indicates a tendency towards memorization, necessitating robust testing and validation protocols. This issue is particularly critical for companies' internal data and tasks that a foundation model has not encountered before, as public benchmarks do not necessarily predict performance on these internal tasks. In addition, it is worth noting that some applications have stringent data privacy policies, a concern increasingly being addressed in the literature \cite{ye_dataframe_2024,carey2024dp,boudewijn2023privacy}. In our work, we assume that the data resides within a secure local network, and we do not address data privacy issues in this paper. In industrial settings, practical constraints such as interpretability, user-centric adaptation, ease of development and maintenance, latency requirements, and IT infrastructure limitations are crucial. Our objective is to design a system that addresses these industrial needs without unnecessary complexity and excessive resources typically required by pre-training and fine-tuning LLMs. 


\section{Conclusion}
\label{sec:conclusion}

We present the GoNoGo, an LLM-based multi-agent system designed to streamline software release decisions in the automotive industry by analyzing and deriving insights from real-world data using Python code. We have employed this system within our industrial partner's company, which is significantly assisting release managers and reducing the number of engineers engaged in this process, allowing them to focus on their high-level tasks.

The impact of our system extends beyond automation, transforming how automotive companies manage their software release cycles. It reduces the time and effort required for data analysis while increasing decision accuracy and reliability. This shift allows engineers and managers to focus on higher-level tasks, accelerating the overall development and deployment process by bridging the gap between raw data and actionable insights, driving the industry towards more efficient, data-driven software release practices. Without GoNoGo in place, our industrial partner would experience more wasted time and effort across various teams and employees, with the decision-making process becoming significantly prolonged. Pilot users have reported saving approximately 2 hours per person each time they make a decision, highlighting the system's positive impact on efficiency and the industrial partner's overall business goals.

\section{Acknowledgement}
This work was partially supported by the Wallenberg AI, Autonomous Sys- tems and Software Program (WASP) funded by the Knut and Alice Wallenberg Foundation. 
%
%
 \bibliographystyle{splncs04}
 \bibliography{biblio}
%
%
%
%
%
\end{document}